\documentclass[sigconf]{acmart-me}

\usepackage{booktabs} 
\usepackage{url}
\usepackage{color}
\usepackage{enumitem}
\usepackage{todonotes}
\setcopyright{rightsretained}

\acmDOI{}

\acmISBN{}

\acmConference[MediaEval'21]{Multimedia Evaluation Workshop}{December 13-15 2021}{Online} 
\acmYear{2021}
\copyrightyear{}

\acmPrice{}

\begin{document}
\title{Overview of The MediaEval 2021 Predicting Media Memorability Task}

\author{Rukiye Savran Kiziltepe\textsuperscript{1}, 
Mihai Gabriel Constantin\textsuperscript{2}, 
Claire-H\'el\`ene Demarty\textsuperscript{3}, 
Graham Healy\textsuperscript{4}, 
Camilo Fosco\textsuperscript{5}, 
Alba G. Seco de Herrera\textsuperscript{1}, 
Sebastian Halder\textsuperscript{1},
Bogdan Ionescu\textsuperscript{2}, 
Ana Matran-Fernandez\textsuperscript{1}, 
Alan F. Smeaton\textsuperscript{4}, 
Lorin Sweeney\textsuperscript{4}}
\affiliation{\textsuperscript{1}University of Essex, UK\\ \textsuperscript{2}University Politehnica of Bucharest, Romania\\ \textsuperscript{3}InterDigital, France\\ \textsuperscript{4}Dublin City University, Ireland\\ \textsuperscript{5}Massachusetts Institute of Technology Cambridge, Massachusetts, USA.}
\email{alba.garcia@essex.ac.uk}

\renewcommand{\shortauthors}{Savran Kiziltepe et al.}
\renewcommand{\shorttitle}{Predicting Media Memorability}

\begin{abstract}
This paper describes the MediaEval 2021 \textit{Predicting Media Memorability} task, which is in its 4th edition this year, as the prediction of short-term and long-term video memorability remains a challenging task. In 2021, two datasets of videos are used: first, a subset of the TRECVid 2019 Video-to-Text dataset; second, the Memento10K dataset in order to provide opportunities to explore cross-dataset generalisation. In addition, an Electroencephalography (EEG)-based prediction pilot subtask is introduced.
In this paper, we outline the main aspects of the task and describe the datasets, evaluation metrics, and requirements for participants' submissions.
\end{abstract}
\maketitle

\section{Introduction}
\label{sc:intro}
%
Information retrieval and recommendation systems deal with exponential growth in media platforms such as social networks and media marketing. New methods of organising and retrieving digital material are needed in order to increase the usefulness of multimedia events in our daily lives. Memorability, like other important video properties, such as aesthetics or interestingness, can be viewed as useful to contribute in the selection of competing videos especially when developing advertising or instructional material. In advertising, predicting the memorability of a video is important since  multimedia materials have varying effects on human memory. In addition to advertising, this task may have an impact on other fields such as film making, education, and content retrieval.

The \textit{Predicting Media Memorability} task addresses this problem. The task is part of the MediaEval benchmark and, following the success of previous editions~\cite{CDD2018,CID2019,Herrera2020Overview,2020Data}, creates a common benchmarking protocol and provides a ground truth dataset for short-term and long-term memorability using common definitions. 

\section{Related Work}
\label{sc:work}
%
The computational study of video memorability is a natural extension of  research into image memorability prediction, which has gained increasing attention in the years after Isola et al.'s work~\cite{IXP2013}. Models have reached remarkable predictive accuracy for image memorability~\cite{KRT2015,SDG2018}, and we have just begun to see the application of approaches such as style transfer to enhance image memorability~\cite{10.1145/3311781}, demonstrating that we have progressed from simply measuring memorability to using it as an evaluation aspect.

In contrast, computer science research on visual memorability (VM) is still in its early stages. Recent studies on video memorability have focused on the short-term~\cite{10.1007/978-3-030-58517-4_14}, but the lack of studies on VM can be explained by a number of factors. To begin with, there are currently not enough publicly available data sets for training and testing models.
The second problem is the absence of a standardised definition for VM. In terms of modeling, previous attempts at VM prediction~\cite{SSS2017,CDD2019} have identified several features that contribute to VM prediction, including semantic, saliency, and colour features. However, the work is far from complete, and our ability to propose effective computational models will aid in meeting the challenge of VM prediction.

The purpose of this task is to contribute to the harmonisation and advancement of this rapidly growing multimedia field. Additionally, in contrast to prior work on image memorability prediction in which memorability was tested only a few minutes after memorisation, we present a dataset containing long-term memorability annotations. We expect that  models trained on this  will produce predictions that are more indicative of long-term memorability, which is preferred in a wide variety of applications. This year we also distribute an external dataset for generalisation purposes and propose a new pilot task which is EEG-based video memorability.

\section{Task Description}
\label{sc:task}
%
The \textit{Predicting Media Memorability} task asks participants to develop automatic systems that predict short-term and long-term memorability scores from short videos. Participants were given a dataset of short videos with short-term and long-term memorability scores, raw annotations, and extracted features. 
Participants were assigned three sub-tasks:
\begin{itemize}
    \item \textbf{Video-based prediction}: Participants are required to generate automatic systems that predict short-term and long-term memorability scores of new videos based on the given video dataset and their memorability scores.
    \item \textbf{Generalization (optional)}: Participants will train their system on one of the two sources of data we provide and will test them on the other source of data. This is an optional sub-task.
    \item \textbf{Electroencephalography (EEG)-based prediction (pilot)}: Participants are required to generate automatic systems that predict short-term memorability scores of new videos based on the given EEG data. This is a pilot sub-task and details for it can be found in~\cite{EEGPilot2021}. 
\end{itemize}

\section{Collection}
\label{sc:collection}
%
This task utilises a subset of the TRECVID 2019 Video-to-Text video dataset~\cite{ABC2019}. The dataset contains Twitter Vine videos where various actions are performed. This year, the dataset has been expanded and normalised short-term memorability scores are provided with memory alpha decay values. Additionally, we  open the Memento10K~\cite{10.1007/978-3-030-58517-4_14} dataset to participants. Apart from traditional video information like metadata and extracted visual features, part of the data will be accompanied by Electroencephalography (EEG) recordings that would allow to explore the physical reactions of  users. 

A set of pre-extracted features are also distributed as follows:
\begin{itemize}
    \item image-level features: AlexNetFC7~\cite{krizhevsky2012imagenet}, HOG~\cite{dalal2005histograms}, HSVHist, RGBHist, LBP~\cite{he1990texture}, VGGFC7~\cite{simonyan2014very}, DenseNet121~\cite{huang2017densely}, ResNet50~\cite{he2016deep}, EfficientNet b3~\cite{tan2019efficientnet};
    \item video-level feature: C3D~\cite{tran2015learning};
    \item audio-level feature: VGGish \cite{hershey2017cnn}.
\end{itemize}
\noindent 

Three frames from each video were used to extract image-level features: the first, the middle, and the last frame. Additionally, each TRECVid video includes at least two textual captions summarising the action, whereas Memento10K includes five. The annotations acquired from participants included the first and second appearance positions of each target video, as well as participants' response times and the keys pressed while watching each video.

\subsection{TRECVid 2019 Video-to-Text dataset}
The TRECVid 2019 Video-to-Text dataset~\cite{ABC2019} contains 6,000 videos. In 2021, three subsets were distributed as part of the MediaEval Predicting Media Memorability task.
The training set contained 588 videos, the development set 1,116 videos and the test set 500 videos. Each video has two associated  memorability scores indicating its likelihood of being remembered after two distinct periods of memory retention. Similar to previous editions of the task~\cite{CDD2018,CID2019}, memorability was measured twice using recognition test: a few minutes after the videos were shown (short-term) and 24-72 hours later (long-term).
The videos are released under Creative Commons licences that allow their redistribution. 
%

The ground truth dataset was generated  using a video memorability game protocol proposed by Cohendet et al.~\cite{CDD2019}. The memorability game was formed in two versions. One was made available on Amazon Mechanical Turk (AMT), and another was made available for general use in three languages: English, Spanish, and Turkish.

In the video memorability game protocol, participants were expected to watch 180 and 120 videos in short-term and long-term memorisation steps, respectively. The goal was essentially to press the space bar whenever participants recognise a previously seen video, which allows for the determination of which videos they do and do not recognise. The game begins with the repetition of 40 target videos after a few minutes to accumulate short-term memorability labels. Regarding the first step's filler videos, 60 non-vigilance filler videos are shown once. After a few seconds, 20 vigilance filler videos are repeated to ensure that participants are paying attention to the task. After 24 to 72 hours, the same individuals are anticipated to return for the second step, which involves collecting labels for long-term memorability. This time, 40 target videos chosen at random from the non-vigilance fillers in the first stage and 80 fillers chosen at random from new videos are displayed to determine the target videos' long-term memorability scores. The percentage of correct recognition for each video is used to calculate short-term and long-term memorability scores. 

\subsection{Memento10K dataset}
The Memento10K dataset~\cite{10.1007/978-3-030-58517-4_14} contains 10,000 three-second videos depicting in-the-wild scenes, with their associated short-term memorability scores, memorability decay values, action labels, and 5 accompanying captions. The scores were computed with 90 annotations per video on average, and the videos were shown to participants without sound. 7,000 videos were released as a training set, and 1,500 were provided for validation. The last 1,500 videos were used as the test set for scoring submissions. 

\section{Submission and Evaluation}
\label{sc:run}
%
As it is with previous tasks, each team is expected to submit both short-term and long-term memorability predictions. A total of ten runs, five for each, can be submitted for video-based prediction. In addition, participants can submit five runs per optional sub-task 
(generalisation and EEG-based prediction). All information, including given features, ground truth data, video sample titles, features extracted from visual material, and even external data, may be used to build the system. Short-term and long-term annotation memorability runs must be submitted separately and must not include each others. 

Classic evaluation metrics (including Spearman's rank correlation) are used to compare the predicted memorability scores for the videos with the ground truth memorability scores.

\section{Discussion and Outlook}
\label{sc:discussion}
%
In this paper we introduced the 4th edition of the Predicting Media Memorability at the MediaEval 2021 Benchmarking initiative. With this task, a comparative assessment of current state-of-the-art machine learning techniques to predict short- and long-term memorability can be conducted. A dataset containing short videos is distributed with memorability annotations and external data is provided for generalisation purposes. Moreover, EEG annotations are also provided for a pilot study. Related information has also been made available to participants so they can refine their strategies.
The 2021 MediaEval workshop proceedings
presents details on the participants' approaches to the task including methodologies used and findings.
\begin{acks}
MGC and BI’s contribution is supported under project AI4Media, a European Excellence Centre for Media, Society and Democracy, H2020 ICT-48-2020, grant \#951911. The work of RSK  is partially funded by the Turkish Ministry of National Education. 
This work was part-funded by NIST Award No. 60NANB19D155 and by Science Foundation Ireland under grant number SFI/12/RC/2289\_P2.
\end{acks}

\bibliographystyle{ACM-Reference-Format}
\def\bibfont{\small} 
\bibliography{references} 

\end{document}